\documentclass[11pt,a4paper]{article}
\usepackage[hyperref]{acl2019}
\usepackage{times}
\usepackage{latexsym}
\usepackage{microtype}
\usepackage{url}

\usepackage{booktabs}

\usepackage{amsmath}
\usepackage{amssymb}

\aclfinalcopy 
\def\aclpaperid{1045} 

\usepackage[safe]{tipa}

\usepackage[disable]{todonotes} 

\newcommand{\saveforCR}[1]{}

\renewcommand{\bold}[1]{{\boldsymbol #1}}

\title{Uncovering Probabilistic Implications in Typological Knowledge Bases}

\author{Johannes Bjerva$^{\textrm{\normalfont\textschwa}}$ \text{ }  Yova Kementchedjhieva$^{\textrm{\normalfont \textschwa}}$ \text{ }  Ryan Cotterell$^{\textrm{\normalfont \textipa{P},\textipa{H}}}$ \text{ } Isabelle Augenstein$^{\textrm{\normalfont \textschwa}}$ \\
${}^{\textrm{\textschwa}}$Department of Computer Science, University of Copenhagen \\
${}^{\textrm{\textipa{P}}}$Department of Computer Science, Johns Hopkins University \\
${}^{\textrm{\textipa{H}}}$Department of Computer Science and Technology, University of Cambridge  \\
{\tt {bjerva,yova,augenstein}@di.ku.dk, rdc42@cam.ac.uk}
}

\date{}

\begin{document}
\maketitle
\begin{abstract}
The study of linguistic typology is rooted in the implications we find between linguistic features, such as the fact that languages with object-verb word ordering tend to have postpositions. 
Uncovering such implications typically amounts to time-consuming manual processing by trained and experienced linguists, which potentially leaves key linguistic universals unexplored. 
In this paper, we present a computational model which successfully identifies known universals, including Greenberg universals, but also uncovers new ones, worthy of further linguistic investigation.
Our approach outperforms baselines previously used for this problem, as well as a strong baseline from knowledge base population.
\end{abstract}

\section{Introduction}

Linguistic typology is concerned with mapping out the relationships between languages with reference to structural and functional properties \citep{crofttypology}.
A typologist may ask, for instance, how a language encodes syntactic features and relationships.
Does it place its verbs before objects or after, and does it have prepositions or postpositions?
It is well established that many features of languages are highly correlated, sometimes to the extent that they imply each other. 
Based on this observation, \citet{greenberg} establishes the notion of implicational universals, i.e., cases where the presence of one feature \textit{strictly} implies the presence of another.

Universals are important to investigate as they offer insight into the inner workings of language and define the space of plausible languages.
Universals can aid cognitive scientists examining the underlying processes of language, as there arguably is a cognitive reason for why, e.g., languages with OV ordering are postpositional \citep{greenberg}.
In the context of natural language processing (NLP), when creating synthetic data for multilingual NLP, one should consider universals to maintain the plausibility of the data \citep{wang}.
Computational typology can furthermore be used to induce language representations, useful in, e.g., language modelling \citep{ostling_tiedemann} and syntactic parsing \citep{delhoneux}. 


\begin{figure}[t]
    \centering
    \includegraphics[width=0.7\columnwidth]{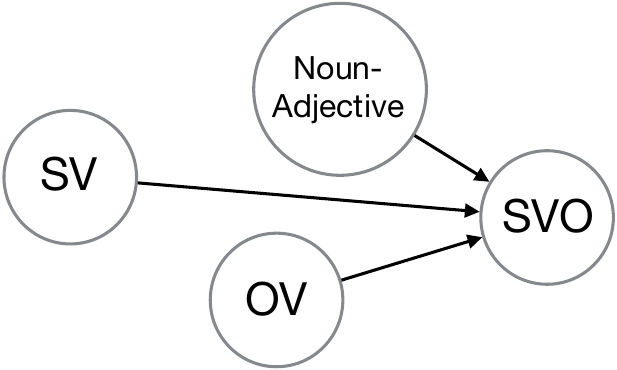}
    \caption{Visualisation of a section of our induced graphical model. Observing the features in the left-most nodes (SV, OV, and Noun-Adjective), can we correctly infer the value of the right-most node (SVO)?}
    \label{fig:belief_net}
\end{figure}
In this paper, we argue that the deterministic Greenbergian view of implications \citep{greenberg} is outdated.
Instead, we suggest that a probabilistic view of implications is more suitable, and define the notion of a probabilistic typological implication as a certain conditional probability distribution.
We do this by first placing a joint distribution over the vector of typological features, and then marginalising out all features other than
the two under consideration. This computation is made tractable by learning a tree-structured graphical model (Figure~\ref{fig:belief_net})
with the PC algorithm of \newcite{neapolitan} and then applying the belief propagation (BP) algorithm \citep{bp}. 
We draw inspiration from manual linguistic efforts to this problem \citep{greenberg,lehmann}, as well as from previous computational methods \citep{daume:2009,bjerva-etal-2019-probabilistic}. 
Additionally, we provide a qualitative analysis of predicted implications, as well as performing an empirical evaluation on typological feature prediction, comparing to strong baselines.

\section{From A Generative Model to Probabilistic Implications}

\paragraph{Notation.}
We now seek a probabilistic formalisation of typological implications.
First, we will introduce the relevant notation.
Let $\ell$ be a language. We will seek to explain the observed, language-specific binary vector of typological features, or parameters, $\bold{\pi}^\ell$ where $\pi^\ell_i = 1$ indicates that the $i^\text{th}$ typological feature is ``on''
in language $\ell$. When it is unambiguous, we will drop the superscript $\ell$. Note that we call the vector $\bold{\pi}$ due to a spiritual similarity to the principle-and-parameters framework of \newcite{chomsky1981lectures}.

\paragraph{A Generative Model of Typology.}
We construct a simple generative probability model over the 
the vector of typological features ${\boldsymbol \pi}$, which factorises
according to some tree structure ${\cal T}$. We will
discuss the provenance of ${\cal T}$ below.
Concretely, this distribution is defined as
\begin{equation}
p({\boldsymbol \pi}) = \prod_{i=1}^{|\bold{\pi}|} p(\pi_i \mid \textit{pa}_{{\cal T}}[\pi_i])
\end{equation}
where $\textit{pa}_{{\cal T}}[\cdot]$ is a function that returns the parents of $\pi_i$, if any, in the tree ${\cal T}$. 
Each conditional $p(\pi_i \mid \textit{pa}_{{\cal T}}[\pi_i])$ 
is treated tabularly with one parameter per table entry: each table entry is a unique configuration of the feature $\pi_i$ and its parents $\textit{pa}_{{\cal T}}[\pi_i]$. We place a symmetric Dirichlet prior with concentration parameter $\alpha= 5$, over each of $p(\pi_i \mid \textit{pa}_{{\cal T}}[\pi_i])$'s table entries. This corresponds to add-5 smoothing. 

\paragraph{Probabilistic Implications.}
Although the original Greenbergian view of implications is deterministic, we argue that a probabilistic approach is more suitable. Indeed, logical implications are a special case of conditional probabilities that only take the values $0$ and $1$, rather than values in $[0, 1]$. Specifically, we argue that probabilistic implications should take the form of the following conditional
probability distribution:
\begin{equation}\label{eq:sum}
p(\pi_i \mid \pi_j) = \sum_{{\boldsymbol \pi}'} p(\pi_i, {\bold{\pi}}' \mid \pi_j)
\end{equation}
where ${\boldsymbol \pi}'$ is a subvector that omits the indices $i$ and $j$. In text, our goal is to sum out all possible languages, holding
two typological features, $\pi_i$ and $\pi_j$, fixed. We note that since our model $p$ factorises according to the tree ${\cal T}$, 
this sum may be performed in polynomial time using dynamic programming, specifically the belief propagation algorithm \citep{bp}. 
Note that we contend this improves upon the ideas of \newcite{daume:2009}, who only considered pair-wise interactions of features: Our definition of probabilistic implications \emph{marginalises out} all other features. 

\paragraph{Discovering Probabilistic Implications.}
How can we use a generative model to discover typological 
implications? What we would like to know is how often $p(\pi_i \mid \pi_j)$ is significantly different than $p(\pi_i)$. We note that $p(\pi_i)$ can also be computed with BP. We now reduce
the search for typological implications as asking when the quantity $|p(\pi_i \mid \pi_j) - p(\pi_i)|$ is statistically significantly greater than 0. Given a sufficiently expressive generative model 
$p$, this allows for a richer notion of implication than Greenberg original proposed, as it admits the softer notion of typological influence.

\paragraph{Learning the Structure of $p$.}
There are many ways to learn the tree structure ${\cal T}$, and we choose the PC algorithm of \newcite{neapolitan}. This algorithm works in two steps---first, it learns a skeleton graph from the data (in our case, a typological data base), with \textit{undirected} edges.
Next, it orients these edges so as to form a directed acyclic graph.
Once we have fit this graph so as to represent $p(\bold{\pi})$, we are left with a tractable model we can use to predict held-out typological features and discover typological implications. 

\paragraph{Parameter Estimation.} 
We apply maximum a posteriori (MAP) inference in order to estimate the parameters of our model. 
If all the data were observed, i.e. there were no missing values in WALS, this could be
achieved by counting and normalising across the typological database in question with the previously mentioned Dirichlet prior. 
(The prior simply corresponds to add-$\lambda$ smoothing.)
However, in many cases we do have missing data.
In fact, we almost never observe all the values in WALS.
Thus, we must rely on expectation-maximisation
to perform MAP estimation \cite{dempster1977maximum}. The gist of the algorithm is simple: we compute ``pseudocounts''
for the missing entries using belief propagation, which we smooth as if they had been observed values. Using these pseudocounts, we get a new estimate of the parameters by count-and-divide as in the fully supervised case. We iterate between updating the pseudocounts and performing count-and-divide.  This is a standard technique in the literature. 


\paragraph{Decoding.} 
In section 4, we are interested in predicting typological features given others.
If we wish to predict $\pi_i$ given observed
features for a language $\bold{\pi}_\textit{obs}$, we compute
\begin{align}
\pi^\star_i &= \mathop{\text{argmax}}  p(\pi_i \mid \bold{\pi}_\textit{obs}) \\
            &= \mathop{\text{argmax}} \sum_{ \bold{\pi}_\textit{unobs}} p(\pi_i, \bold{\pi}_\textit{unobs} \mid \bold{\pi}_\textit{obs}) 
\end{align}
where we marginalize out all those features $\bold{\pi}_\textit{unobs}$ unobserved or held out in a given language.  The conditional may be computed with belief propagation and the argmax is over the set $\{0, 1\}$. This makes the computation tractable.





\section{WALS: A Typological Database}

\begin{figure}[tb]
    \centering
    \includegraphics[width=\columnwidth]{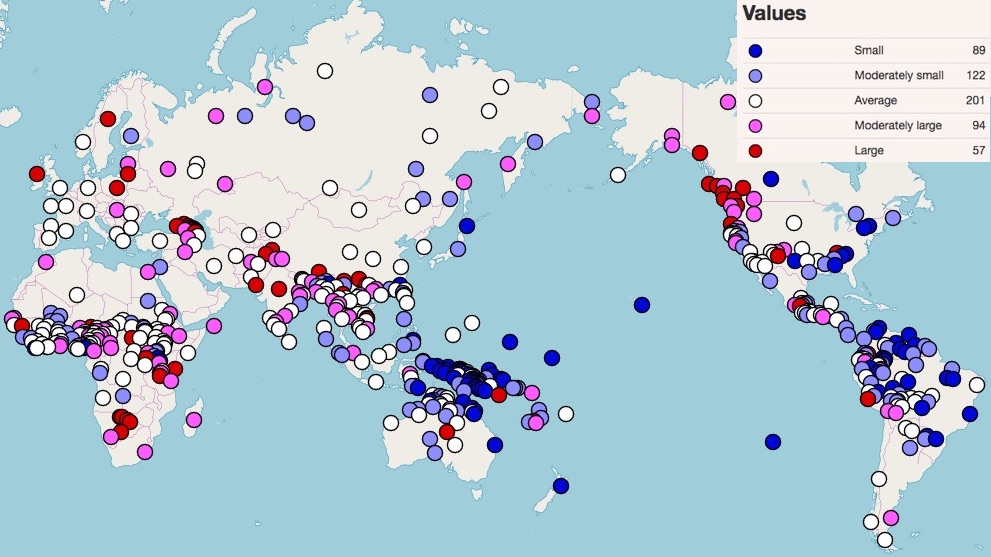}
    \caption{Consonant inventory sizes across languages in the world (WALS, \citet{wals}).}
    \label{fig:wals}
\end{figure}

Before explaining our experimental setup, we first explain the data set we use in evaluation.
We evaluate on the World Atlas of Language Structure (WALS, \citet{wals}, which is the largest openly available typological database. It comprises approximately 200 linguistic features with annotations for more than 2,500 languages.
These annotations have been made by expert typologists through meticulous study of grammars and field work.
WALS is quite sparse, however, as only 100 of these languages have annotations for all features.
For instance, Figure~\ref{fig:wals} shows the distribution of consonant inventory sizes across the languages for which this feature is annotated.
Although this is not our main contribution, the fact that we can predict held-out features offers a way to fill in the feature value gaps which exist for the vast majority of languages.

\paragraph{Pre-processing}
We pre-process our data similarly to \citet{daume:2009}.
We filter out features which are not encoded for at least 100 languages, and feature values which occur for fewer than 10\% of the languages.
The reason for this is that any implications found for exceedingly rare features is likely to be inconclusive.
We further follow \citet{daume:2009} in that we binarise features with more than 7 feature values such that they simply encode whether or not a language has a feature.
For instance, features are not likely to have implicants determining the \textit{number} of tones, but rather the presence or absence of tones.
Finally, they take into account that languages are not independent, as phylogenetic similarity can help infer features in closely related languages.
We do not use this information, as we are interested in finding implications which ought to be independent of language relatedness.

\section{Two Typological Experiments}

In order to evaluate our probabilistic approach to typological implications, we define two tasks.
Our \textbf{empirical evaluation} is based on predicting features so as to get an objective measure of our model, which is comparable both to previous work and other strong baselines.
Second, we include a \textbf{qualitative evaluation}, as we are interested in uncovering both known and novel typological implications.


\begin{table}
    \centering
    \resizebox{\columnwidth}{!}{
    \begin{tabular}{lrrrrr}
            \toprule
            N implicants       & 2 & 3 & 4 & 5 & 6  \\
            \midrule
            Phonology          & 0.75 & 0.82 & 0.84 & 0.86 & \textbf{0.89} \\
            Morphology         & 0.77 & 0.85 & \textbf{0.87} & 0.70 & 0.82 \\
            Nominal Categories & 0.72 & 0.83 & 0.80 & \textbf{0.84} & 0.81 \\
            Nominal Syntax     & 0.77 & \textbf{0.89} & 0.85 & \textbf{0.89} & 0.81 \\
            Verbal Categories  & 0.80 & 0.84 & 0.80 & 0.86 & \textbf{0.90} \\
            Word Order         & 0.74 & 0.86 & 0.86 & 0.86 & \textbf{0.93} \\
            Clause             & 0.75 & 0.81 & 0.84 & \textbf{0.85} & 0.84 \\
            Complex            & 0.82 & 0.83 & 0.87 & \textbf{0.93} & 0.84 \\
            Lexical            & 0.83 & 0.76 & 0.75 & \textbf{0.85} & 0.79 \\
            \midrule
            Mean               & 0.77 & 0.83 & 0.83 & \textbf{0.85} & \textbf{0.85} \\
            \midrule
            \midrule
            Most freq.         & \multicolumn{5}{c}{\textit{0.30}}\\ 
            Pairwise           & \multicolumn{5}{c}{\textit{0.77}}\\ 
            PRA                & \multicolumn{5}{c}{\textit{0.81}}\\
            Language embeddings & \multicolumn{5}{c}{\textit{0.85}}\\
            \bottomrule
    \end{tabular}
    }
    \caption{Accuracies for feature prediction in a typologically diverse test set, across number of implicants used. Note that the numbers are not comparable across columns nor to the baseline, since each makes a different number of predictions.}
    \label{tab:prediction}
\end{table}

\subsection{Predicting Typological Features}

Feature prediction is a commonly used task in evaluating how well a given model is able to explain the typological features of languages
\citep{daume:2009,malaviya,cotterell:vowelform,ctsurvey,bjerva-etal-2019-probabilistic}.
This is an important task which can highlight the extent to which a model has learned interdependencies between languages and features.
We include this evaluation to first show that our model has predictive power which surpasses strong baselines, before investigating the main research question of this work, i.e., the extent to which we can uncover probabilistic implications.
We evaluate the models on feature prediction by fitting our model on 80\% of the languages in WALS, and leaving out 10\% of the languages for development and testing, respectively.

We split our evaluation of our model up across the feature categories present in WALS.
These cover areas such as phonology, morphology etc., listed in Table~\ref{tab:prediction}.
During the typological feature prediction experiments, we consider a single such WALS category at a time.\looseness=-1
We vary the number of implicants by allowing the model to observe 2 to 6 features from within this category as well as the values of features in \emph{other} categories. 
This is done as having access to, e.g., all word-order features when predicting a final word order feature would be much easier than our setting.
Hence, our experiment will show the extent to which increasing the number of features from the current feature category affects predictive power.
We vary the number of implicants $k$ from 2 to 6 features in each category with a total of $n$ features, this gives us $\binom{n}{k}$ total sets per number of implicants $k$. For each
such set, we attempt to predict all held-out
features in that category in a leave-one-out-style evaluation. This results in $\binom{n}{k} (n-k)$ predictions to make per category per number of implicants $k$. 
Performance is measured by averaging the accuracy of predictions of all held-out features over the entire test set, across categories.

\paragraph{Baseline \#1: Most frequent}
Since many typological features have low-entropy distributions, a most frequent class baseline is a relatively strong lower bound for prediction of typological features.
For instance, this yields an accuracy of 45\% when predicting the canonical subject--object--verb ordering in a language.

\paragraph{Baseline \#2: Pairwise prediction}
We implement a simple baseline based on pairwise prediction of typological features.
This is inspired by the approach in \citet{daume:2009}. 
As this code was not publicly available we provide our own non-Bayesian implementation.

\paragraph{Baseline \#3: PRA} Since WALS can be seen as a knowledge base, we apply a strong baseline from the field of knowledge base population.
Path Ranking Algorithm (PRA) is an algorithm which finds relation paths by traversing the knowledge graph, which can then be used to predict implicatures and feature values \citep{Lao:2010:RRU:1842816.1842823,Lao:2011:RWI:2145432.2145494}.\footnote{We use the original implementation of PRA available here: \url{https://github.com/noon99jaki/pra}}
We train PRA using the standard hyperparameters of the existing implementation, which includes regularising with $\ell_1=0.001$ and $\ell_2=0.001$, as well as using negative sampling.

\paragraph{Baseline \#4: Language embeddings} Although we aim to predict \textit{implications}, and not only feature values, we compare with previous work on predicting typological features in WALS \citep{bjervaaugenstein:2018}.
As their setup is different, we use their highest reported score as a baseline.

\paragraph{Feature Prediction Results.}
Table~\ref{tab:prediction} contains the results from feature prediction across the chapters outlined in WALS.
Our implementation is able to predict features across categories above baseline levels.
At increasing numbers of implicants, prediction power tends to increase.
This is not the case for all feature categories, however.
One such case is Nominal Syntax, in which performance peaks at 3 implicants.
This is expected, as correlations only exist between some features, thus at a certain point access to more typological features no longer helps performance.
Note that although the baseline numbers are based on predicting the same features as our model, the baseline models do not observe the same features during prediction - for instance Baseline \#4 does not make predictions based on other feature values, but is trained on one feature at a time.

\subsection{Discovering Typological Implications}

\newcounter{rtaskno}
\newcommand{\rtask}[1]{\refstepcounter{rtaskno}\label{#1}}
\begin{table}
    \centering
  \resizebox{\columnwidth}{!}{
    \begin{tabular}{lr@{\hskip3pt}c@{\hskip3pt}l}
    \toprule
        \# & Implicant & $\supset$ & Implicand  \\
        \midrule
        \ref{implication:post_gen}* & Postpositions & $\supset$ & Genitive-Noun (Greenberg \#2a) \rtask{implication:post_gen}\\
        \ref{implication:post_OV}* & Postpositions & $\supset$ & OV    (Greenberg \#4)  \rtask{implication:post_OV} \\
        \ref{implication:ov_sv}  & OV & $\supset$ & SV    \rtask{implication:ov_sv}\\
        \ref{implication:post_sv}* & Postpositions & $\supset$ & SV   \rtask{implication:post_sv}\\
        \ref{implication:prep_vo}* & Prepositions & $\supset$ & VO    (Greenberg \#4) \rtask{implication:prep_vo} \\
        \ref{implication:prep_init}* & Prepositions & $\supset$ & Initial subord. word (Lehmann) \rtask{implication:prep_init} \\
        \midrule
        \ref{implication:adjpost_demn}* & Adjective-Noun, Postpositions & $\supset$ & Demonstrative-Noun  \rtask{implication:adjpost_demn}\\
        \ref{implication:gennadjn_ov}* & Genitive-Noun, Adjective-Noun & $\supset$ & OV  \rtask{implication:gennadjn_ov} \\
        \midrule
        
         

        \ref{implication:svov_SOV} & SV && \\ & OV && \\ & Noun-Adjective & $\supset$ & SOV \rtask{implication:svov_SOV} \\

        \ref{implication:numeralnoun} &  Degree word-Adjective && \\ & VO and Noun--Relative Clause && \\ & SVO & $\supset$ & Numeral-Noun \rtask{implication:numeralnoun} \\
        \ref{implication:nounnumeral} &  SOV && \\ & OV and Relative Clause--Noun && \\ & Adjective-Degree word & $\supset$ & Noun-Numeral \rtask{implication:nounnumeral}\\
        
        
        
        
    \bottomrule
         
    \end{tabular}
    }
    \caption{Hand-picked implications. In cases where the same is covered by \citet{daume:2009}, we borrow their analysis (marked with *).}
    \label{tab:implications}

\end{table}


Having established that our method bests several competitive baselines for prediction of typological features, we next look at what implications our probabilisation of typology allows us to find. 
We search for those conditional probabilities where the quantity $|p(\pi_i \mid \pi_j) - p(\pi_i)|$ is statistically significantly greater than 0, as found with an independent two-tailed t-test.\footnote{Future work will make use of a non-parametric test, whose details
we are still working out.}
After adjusting for multiple tests with the Bonferroni correction, we report those implications where $p < 0.05$.
We report the full list of implications found by our model in the Supplements and show a subset of these in  Table~\ref{tab:implications}.\footnote{Also on \url{bjerva.github.io/imp_acl19.pdf}.}
We note that we are able to find the same implications listed by \citet{daume:2009}, some of which are listed in the table.
These implications include Greenberg universals \citep{greenberg}, showing that our approach to probabilisation of linguistic universals is suitable to replicate previous work.



\paragraph{Transitivity across implications}
At first glance, it is not clear why postpositions should imply SV word order, as stated in \#\ref{implication:post_sv}. 
Yet, \#\ref{implication:post_OV} is a well-established universal \citep{greenberg} and \#\ref{implication:ov_sv} comes with strong statistical evidence: SV order is much more frequent than VS word order in OV languages (98.44\% of these are predominantly SV). Our model has thus used transitive reasoning of the form \textit{if $ A \supset B \wedge B \supset C$ then $A \supset C$} to find \#\ref{implication:post_OV}. 


\paragraph{The power of multiple implicants}
Implications \#\ref{implication:numeralnoun} and \#\ref{implication:nounnumeral} concern the order between nouns and their numeral modifiers. The two main alternatives here, Noun-Numeral and Numeral-Noun are of comparable frequency in WALS; they occur in 607 and 479 languages, respectively, i.e.~Noun-Numeral holds the majority with only 55\%. If we consider each of the three implicants listed in implication \#\ref{implication:nounnumeral} on their own, the strongest statistical power goes to the \textit{Degree word--Adjective} feature: conditioned on this feature, the Numeral-Noun order holds in 79\% of the relevant languages. The combination of all three implicants, on the other hand, results in a subset of languages with 91\% Numeral-Noun order. The Numeral-Noun order can thus be implied with considerably more confidence from a combination of multiple implicants.

\section{Related Work}
Typological implications outline the space of possible languages, based on evidence from observed languages, as recorded and classified by linguists \citep{greenberg,lehmann,hawkins}.
While work in this direction has been manual, typological knowledge bases do exist now \citep{wals, uriel}, which allows for automated discovery of implications.
Although previous computational work exists \citep{daume:2009}, we are the first to introduce a probabilisation of typological implications.

In addition to work on finding implications based on known features, there is an increasing amount of work on computational methods to discovering typological features \citep{ctsurvey}.
Work in this area includes unsupervised discovery of word order \citep{ostling:2015} or other linguistic features \citep{asgari}, typological probing of language representations \citep{bjerva2019language,beinborn2019semantic}, and several papers attempt to predict typological features in WALS \cite{georgi2010,malaviya,bjervaaugenstein:2018,bjerva-augenstein-2018-tracking,cotterell:vowelinv,cotterell:vowelform,bjerva-etal-2019-probabilistic}.

\section{Conclusions}
We defined the notion of probabilistic implications, and presented a computational model which successfully identifies known universals, including Greenberg universals, but also uncovers new ones, worthy of further investigation by typologists.
Additionally, our approach outperforms strong baselines for prediction of typological features.

\section*{Acknowledgments}
We acknowledge the computational resources provided by CSC in Helsinki through NeIC-NLPL (www.nlpl.eu), and the support of the NVIDIA Corporation with the donation of the Titan Xp GPU used for this research. 

\bibliographystyle{acl_natbib}
\bibliography{tacl}

\end{document}